\documentclass[10pt,journal,cspaper,compsoc]{IEEEtran}
\usepackage{times}
\usepackage{epsfig}
\usepackage{graphicx}
\usepackage{amsmath}
\usepackage{amsfonts}
\usepackage{amssymb}
\usepackage{url}
\usepackage{setspace}

\usepackage{caption}
\usepackage{framed,multirow}
\usepackage{latexsym}
\usepackage{subfigure}
\usepackage{algorithm2e}
\usepackage{booktabs}
\usepackage[pagebackref=true,breaklinks=true,letterpaper=true,colorlinks,bookmarks=false]{hyperref}

\ifCLASSOPTIONcompsoc
\else
\fi
\ifCLASSINFOpdf
\else
\fi
\begin{document}
%
\title{A Robust 3D-2D Interactive Tool for Scene Segmentation and Annotation}
%
%
%
%

\author{Duc~Thanh~Nguyen, Binh-Son Hua$^*$\thanks{$^*$Co-first author}, Lap-Fai Yu,~\IEEEmembership{Member,~IEEE,} and Sai-Kit Yeung,~\IEEEmembership{Member,~IEEE}
\IEEEcompsocitemizethanks{\IEEEcompsocthanksitem Duc Thanh Nguyen is with the School of Information Technology, Deakin University, Australia. This work was done when Duc Thanh Nguyen was working at the Singapore University of Technology and Design.\protect\\
E-mails: duc.nguyen@deakin.edu.au
\IEEEcompsocthanksitem Lap-Fai Yu is with the University of Massachusetts Boston.\protect\\
E-mail: craigyu@cs.umb.edu
\IEEEcompsocthanksitem Sai-Kit Yeung is with Information Systems Technology and Design Pillar, Singapore University of Technology and Design.\protect\\
E-mail: saikit@sutd.edu.sg}}
\IEEEcompsoctitleabstractindextext{%
\begin{abstract}

Recent advances of 3D acquisition devices have enabled large-scale acquisition of 3D scene data. Such data, if completely and well annotated, can serve as useful ingredients for a wide spectrum of computer vision and graphics works such as data-driven modeling and scene understanding, object detection and recognition. However, annotating a vast amount of 3D scene data remains challenging due to the lack of an effective tool and/or the complexity of 3D scenes (e.g. clutter, varying illumination conditions). This paper aims to build a robust annotation tool that effectively and conveniently enables the segmentation and annotation of massive 3D data. Our tool works by coupling 2D and 3D information via an interactive framework, through which users can provide high-level semantic annotation for objects. We have experimented our tool and found that a typical indoor scene could be well segmented and annotated in less than 30 minutes by using the tool, as opposed to a few hours if done manually. Along with the tool, we created a dataset of over a hundred 3D scenes associated with complete annotations using our tool. The tool and dataset are available at \url{www.scenenn.net}.

\end{abstract}

}

\maketitle

\IEEEdisplaynotcompsoctitleabstractindextext

%
\IEEEpeerreviewmaketitle

\section{Introduction}
\label{intro}

\def\totalscenes{100}

\begin{figure*}[t]
    \centering
    \includegraphics[page=1]{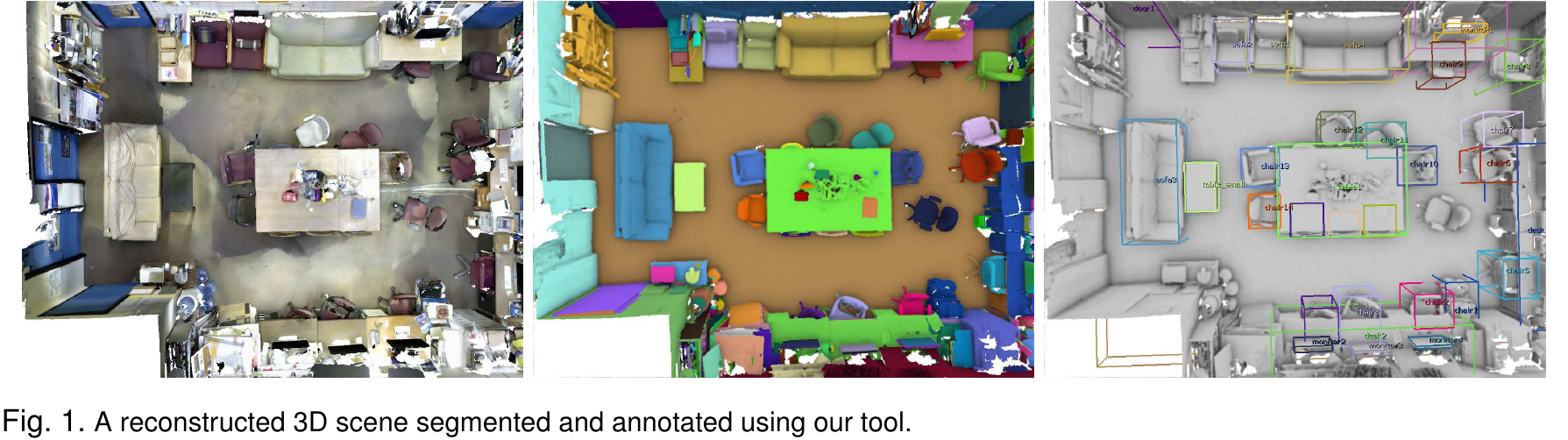}
    \caption{}
    \label{fig:teaser}
\end{figure*}

\IEEEPARstart{H}{igh}-quality 3D scene data has become increasingly available thanks to the growing popularity of consumer-grade depth sensors and tremendous progress in 3D scene reconstruction research \cite{Roth_BMVC_2012}, \cite{Shotton_CVPR_2013}, \cite{Xiao_ICCV_2013}, \cite{Zhou_CVPR_2014}. Such 3D data, if fully and well annotated, would be useful for powering different computer vision and graphics tasks such as scene understanding \cite{Valentin_CVPR_2013}, \cite{Hane_CVPR_2013}, object detection and recognition \cite{Wu_CVPR_2015}, and functionality reasoning in 3D space \cite{Gupta_CVPR_2011}.

Scene segmentation and annotation refer to separating an input scene into meaningful objects. For example, the scene in Fig.~\ref{fig:teaser} can be segmented and annotated into chairs, table, etc. Literature has shown the crucial role of 2D annotation tools (e.g. \cite{Russell_IJCV_2008}) and 2D image datasets (e.g. \cite{Deng_CVPR_2009}, \cite{Everingham_IJCV_2010}, \cite{Xiao_CVPR_2010}) in various computer vision problems such as semantic segmentation, object detection and recognition \cite{Torralba_PAMI_2008}, \cite{Deng_ECCV_2010}. This inspires us for such tasks on 3D scene data. However, segmentation and annotation of 3D scenes require much more effort due to the large scale of the 3D data (e.g. there are millions of 3D points in a reconstructed scene). Development of a robust tool to facilitate the segmentation and annotation of 3D scenes thus is a demand and also the aim of this work. To this end, we make the following contributions:
\begin{itemize}
  \item[$\bullet$] We propose an interactive framework that effectively couples the geometric and appearance information from multi-view RGB data. The framework is able to automatically perform 3D scene segmentation.
  \item[$\bullet$] Our tool is facilitated with a 2D segmentation algorithm based on 3D segmentation.
  \item[$\bullet$] We develop assistive user-interactive operations that allow users to flexibly manipulate scenes and objects in both 3D and 2D. Users co-operate with the tool by refining the segmentation and providing semantic annotation.
  \item[$\bullet$] To further assist users in annotation, we propose an object search algorithm which automatically segments and annotates repetitive objects defined by users.
  \item[$\bullet$] We create a dataset with more than hundred scenes. All the scenes are fully segmented and annotated using our tool. We refer readers to~\cite{Hua_Scenenn_3DV_2016} for more details and proof-of-concept applications using the dataset. 
\end{itemize}

Compared with existing works on RGB-D segmentation and annotation (e.g. \cite{Ren_CVPR_2012}, \cite{Gupta_CVPR_2013}), our tool offers several advantages. First, in our tool, segmentation and annotation are centralized in 3D which free users from manipulating thousands of images. Second, the tool can adapt with either RGB-D images or the triangular mesh of a scene as the input. This enables the tool to handle meshes reconstructed from either RGB-D images \cite{Choi_CVPR_2015} or structure-from-motion \cite{Jancosek_CVPR_2011} in a unified framework.

We note that interactive annotation has also been exploited in a few concurrent works, e.g. SemanticPaint in \cite{Valentin_TOG_2015} and Semantic Paintbrush in \cite{Miksik_CHI_2015}. However, those systems can only handle scenes that are partially captured at hand and contain a few of objects to be annotated. In contrast, our annotation tool handles complete 3D scenes and is able to work with pre-captured data. Our collected scenes are more complex with a variety of objects. Moreover, the SemanticPaint \cite{Valentin_TOG_2015} requires physical touching for the interaction and hence limits its capability to touchable objects. Meanwhile, objects at different scales can be annotated using our tool. In addition, the tool also supports 2D segmentation which is not available in both SemanticPaint \cite{Valentin_TOG_2015} and Semantic Paintbrush \cite{Miksik_CHI_2015}.

\section{Related Work}
\label{sec:relatedwork}

\textbf{RGB-D Segmentation.} A common approach for scene segmentation is to perform the segmentation on RGB-D images and use object classifiers for labeling the segmentation results. Examples of this approach can be found in \cite{Ren_CVPR_2012}, \cite{Gupta_CVPR_2013}. The spatial relationships between objects can also be exploited to infer the scene labels. For example, Jia et al. \cite{Jia_CVPR_2013} used object layout rules for scene labeling. The spatial relationship between objects was modeled by a conditional random field (CRF) in \cite{Lin_ICCV_2013, Kim_ICCV_2013} and directed graph in \cite{Wong_CGF_2015}.

In general, the above methods make use of RGB-D images captured from a single viewpoint of a 3D scene and thus could partially annotate the scene. Compared with those methods, our tool can achieve more complete segmentation results with the 3D models of the scene and its objects.\\

\noindent\textbf{From 2D to 3D Labeling.} Compared with 2D labels, 3D labels are often desired as they provide a more comprehensive understanding of the real world. 3D labels can be propagated by back-projecting 2D labels from image domain to 3D space. For example, Wang et al. \cite{Wang_CVPR_2013} used the labels provided in the ImageNet \cite{Deng_CVPR_2009} to infer 3D labels. In \cite{Xiao_ICCV_2013}, 2D labels were obtained by drawing polygons.

Labeling directly on images is time consuming. Typically, a few thousands of images need to be handled. It is possible to perform matching among the images to propagate the annotations from one image to another, e.g. \cite{Xiao_ICCV_2013}, but this process is less reliable.\\


\begin{figure*}[t]
    \centering
    \includegraphics[page=2]{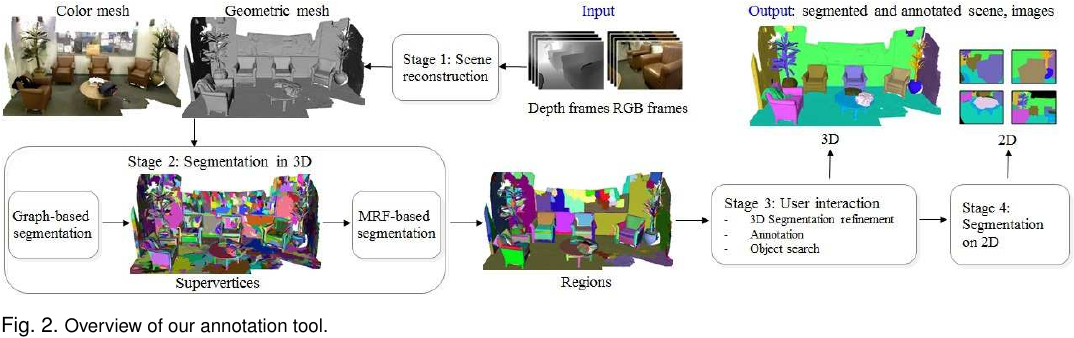}
    \caption{}
    \label{fig:flowchart}
\end{figure*}

\noindent\textbf{3D Object Templates.} 3D object templates can be used to segment 3D scenes. The templates can be organized in holistic models, e.g., \cite{Kim_TOG_2012}, \cite{Salas-Moreno_CVPR_2013}, \cite{Nan_TOG_2012}, \cite{Shao_TOG_2012}, or part-based models, e.g. \cite{Chen_TOG_2014}. The segmentation can be performed on 3D point clouds, e.g. \cite{Kim_TOG_2012}, \cite{Nan_TOG_2012}, \cite{Chen_TOG_2014}, or 3D patches, e.g. \cite{Shao_TOG_2012}, \cite{Salas-Moreno_CVPR_2013}, \cite{Zhang_TOG_2015}.

Generally speaking, the above techniques require the template models to be known in advance. They do not fit well our interactive system in which the templates can be provided on the fly by users. In our tool, we propose to use shape matching to help users in the segmentation and annotation task. Shape matching does not require off-line training and is proved to perform efficiently in practice.\\

\noindent\textbf{Online Scene Understanding.} Recently, there are methods that directly combine 3D reconstruction with annotation to achieve online scene understanding. For example, SemanticPaint proposed in \cite{Valentin_TOG_2015} allowed users annotate a scene by touching objects of interest. A CRF was then constructed to model each indicated object and then used to parse the scene. The SemanticPaint was extended to the Semantic Paintbrush in \cite{Miksik_CHI_2015} for outdoor scenes annotation by exploiting the farther range of a stereo rig.

In both \cite{Valentin_TOG_2015} and \cite{Miksik_CHI_2015}, annotated objects and user-specified objects are assumed to have similar appearance (e.g. color). Furthermore, since the CRF models are built upon the reconstructed data, it is implicitly assumed that the reconstructed data is good enough so that the CRF model constructed from the user-specified object and that of the objects to be annotated have consistent geometric representation. However, the point cloud of the scene is often incomplete, e.g. there are holes. To deal with this issue, we describe the geometric shape of 3D objects using a shape descriptor which is robust to shape variation and occlusion. Experimental results show that our approach works well under noisy data (e.g. broken mesh) and robustly deal with shape deformation while being efficient for practical use.

Online interactive labeling is a trend for scene segmentation and annotation in which the scalability and convenience of the user interface are important factors. In \cite{Valentin_TOG_2015}, the annotation can only be done for objects that are physically touchable and hence is limited to partial scenes. In this sense, we believe that our tool would facilitate the creation of large-scale, complete, and semantically annotated 3D scene datasets.

\section{System Overview}
\label{sec:systemoverview}

Fig.~\ref{fig:flowchart} shows the workflow of our tool. The tool includes four main stages: scene reconstruction, automatic 3D segmentation, interactive refinement and annotation, and 2D segmentation.

In the first stage (section~\ref{sec:reconstruction}), the system takes a sequence of RGB-D frames and reconstructs a triangular mesh, called \emph{3D scene mesh}. After the reconstruction, we compute and cache the correspondences between the 3D vertices in the reconstructed scene and the 2D pixels on all input frames. This allows seamless switching between segmentation in 3D and 2D in later steps.

In the second stage (section~\ref{sec:segmentation}), the 3D scene mesh is automatically segmented. We start by clustering the mesh vertices into supervertices (section~\ref{sec:graphcut}). Next, we group the supervertices into regions (section~\ref{sec:MRF}). We also cache the results of both the steps for later use.

The third stage (section~\ref{sec:refinement}) of the system is designed for users to interact with the system. We design three segmentation refinement operations: \emph{merge}, \emph{extract}, and \emph{split}. After refinement, users can make semantic annotation for objects in the scene.

To further assist users in segmentation and annotation of repetitive objects, we propose an algorithm to automatically search for repetitive objects specified by a template (section~\ref{sec:objectsearch}). We extend the well-known 2D shape context \cite{Belongie_PAMI_2002} to 3D space and apply shape matching to implement this functionality.

The fourth stage of the framework (section~\ref{sec:2Dsegmentation}) is designed for segmentation of 2D frames. In this stage, we devise an algorithm that uses the segmentation results in 3D as initiative for the segmentation on 2D and bases on contour matching.


\section{Scene Reconstruction}
\label{sec:reconstruction}

\subsection{Geometry reconstruction}
\label{sec:georeconstruction}

Several techniques have been developed for 3D scene reconstruction. For example, KinectFusion~\cite{Newcombe_ISMAR_2011} applied frame-to-model alignment to fuse depth information and visualize 3D scenes in real time. However, KinectFusion tends to cause drift where depth maps are not accurately aligned due to accumulation of registration errors over time. Several attempts have been made to avoid drift and led to significant improvements in high-quality 3D reconstruction. For example, Xiao et al.~\cite{Xiao_ICCV_2013} added object constraints to correct misaligned reconstructions. Zhou et al.~\cite{Zhou_TOG_2013}, \cite{Zhou_CVPR_2014} split input frames into small chunks, each of which could be accurately reconstructed using a standard SLAM system like KinectFusion. An optimization was then performed to register all the chunks into the same coordinate frame. In robotics, SLAM systems also detect re-visiting places and trigger a loop closure constraint to enforce global consistency of camera poses.

In this work, we adopt the system in~\cite{Zhou_CVPR_2014}, \cite{Choi_CVPR_2015} to calculate camera poses. Given the camera poses, the triangular mesh of a scene can be extracted using the marching cubes algorithm \cite{Roth_BMVC_2012}. We also store the camera pose of each input frame for computing 3D-2D correspondences. The normal of each mesh vertex is given by the area-weighted average over the normals of its neighbor surfaces. We further smooth the resulting normals using a bilateral filter.

\subsection{3D-2D Correspondence}
\label{sec:3D2Dcorrespondence}

Given the reconstructed 3D scene, we align the whole sequence of 2D frames with the 3D scene using the corresponding camera poses obtained from section~\ref{sec:georeconstruction}. For each vertex, the normal is computed directly on the 3D mesh and its color is estimated as the median of the color of the corresponding pixels on 2D frames.

\section{Segmentation in 3D}
\label{sec:segmentation}

After the reconstruction, a scene mesh typically consists of millions of vertices. In this stage, those vertices are segmented into much fewer regions. To achieve this, we first divide the reconstructed scene into a number of so-called supervertices by applying a purely geometry-based segmentation method. We then merge the supervertices into larger regions by considering both surface normals and colors. We keep all the supervertices and regions for later use. In addition, the hierarchical structures of the regions, supervertices, and mesh vertices (e.g. list of mesh vertices composing a supervertex) are also recorded.

\subsection{Graph-based Segmentation}
\label{sec:graphcut}

We extend the efficient graph-based image segmentation algorithm of Felzenszwalb et al. \cite{Felzenszwalb_IJCV_2004} to 3D space. Specifically, the algorithm operates on a graph defined by the scene mesh in which each node in the graph corresponds to a vertex in the mesh. Two nodes in the graph are linked by an edge if their two corresponding vertices in the mesh are the vertices of a triangle. Let $\mathbf{V} = \{\mathbf{v}_i\}$ be the set of vertices in the mesh. The edge connecting two vertices $\mathbf{v}_i$ and $\mathbf{v}_j$ is weighted as
\begin{equation}
  \label{eq:edgeweight}
    w(\mathbf{v}_i, \mathbf{v}_j) = 1 - {{\mathbf{n}}_i}^\top {\mathbf{n}}_j,
\end{equation}
where ${\mathbf{n}}_i$ and ${\mathbf{n}}_j$ are the unit normals of $\mathbf{v}_i$ and $\mathbf{v}_j$ respectively.

The graph-based segmenter in \cite{Felzenszwalb_IJCV_2004} employs a number of parameters including a smoothing factor used for noise filtering (normals in our case), a threshold representing the contrast between adjacent regions, and the minimum size of segmented regions. In our implementation, those parameters were set to 0.5, 500, and 20 respectively. However, we also make those parameters available to users for customization.

The graph-based segmentation algorithm results in a set of supervertices $\mathcal{S} = \{s_i\}$. Each supervertex is a group of geometrically homogeneous vertices with similar surface normals. The bottom left image in Fig.~\ref{fig:flowchart} shows an example of the supervertices. More examples can be found in Fig.~\ref{fig:results1} and Fig.~\ref{fig:results2}.

\subsection{MRF-based Segmentation}
\label{sec:MRF}

The graph-based segmentation often produces a large number (e.g. few thousands) of supervertices which could require considerable effort for annotation. To reduce this burden, the supervertices are clustered into regions via optimizing an MRF model. In particular, for each supervertex $s_i \in \mathcal{S}$, the color and normal of $s_i$, denoted as $\bar{\textbf{c}}_i$ and $\bar{\textbf{n}}_i$, are computed as the means of the color values and normals of all vertices $\mathbf{v} \in s_i$. Each supervertex $s_i \in \mathcal{S}$ is then represented by a node $o_i$ in an MRF. Two nodes $o_i$ and $o_j$ are directly connected if $s_i$ and $s_j$ share some common boundary (i.e. $s_i$ and $s_j$ are adjacent supervertices). Let $l_i$ be the label of $o_i$, the unary potentials are defined as
\begin{equation}
  \label{eq:likelihood}
    \psi_1(o_i, l_i) = -\log \mathcal{G}^c_i(\bar{\textbf{c}}_i, \boldsymbol\mu^c_{l_i}, \boldsymbol\Sigma^c_{l_i}) - \log \mathcal{G}^n_i(\bar{\textbf{n}}_i, \boldsymbol\mu^n_{l_i}, \boldsymbol\Sigma^n_{l_i}),
\end{equation}
where $\mathcal{G}^c_{l_i}$ and $\mathcal{G}^n_{l_i}$ are the Gaussians of the color values and normals of the label class of $l_i$, $\boldsymbol\mu^c_{l_i}/\boldsymbol\mu^n_{l_i}$ and $\boldsymbol\Sigma^c_{l_i}/\boldsymbol\Sigma^n_{l_i}$ are the mean and covariance matrix of $\mathcal{G}^c_{l_i}/\mathcal{G}^n_{l_i}$.

The pairwise potentials are defined as the Potts model \cite{Barker_PR_2000}
\begin{align}
  \label{eq:prior}
    \psi_2(l_i, l_j) = \begin{cases}
                    -1, & \mbox{if } l_i = l_j \\
                    1, & \mbox{otherwise}.
              \end{cases}
\end{align}

Let $\mathcal{L}=\{l_1, l_2, ..., l_{|\mathcal{S}|}\}$ be the set of labels of supervertices. The optimal labels $\mathcal{L}^*$ is determined by
\begin{align}
  \label{eq:energy}
   \mathcal{L}^* = \operatorname*{arg\,min}_{\mathcal{L}} \bigg[ \sum_{i} \psi_1(o_i, l_i) + \gamma \sum_{i,j} \psi_2(l_i, l_j) \bigg]
\end{align}
where $\gamma$ is weight factor set to 0.5 in our implementation.

The optimization problem in (\ref{eq:energy}) is solved using the method in \cite{Barker_PR_2000}. In our implementation, the number of labels was initialized to the number of supervertices; each supervertex was assigned to a different label. Fig.~\ref{fig:flowchart} (bottom) shows the result of the MRF-based segmentation. More results of this step are presented in Fig.~\ref{fig:results1} and Fig.~\ref{fig:results2}.

\section{Segmentation Refinement and Annotation in 3D}
\label{sec:refinement}

The automatic segmentation stage could produce over- and under- segmented regions. To resolve these issues, we design three operations: \emph{merge}, \emph{extract}, and \emph{split}.

\begin{figure}[t]
    \centering
    \includegraphics[page=3]{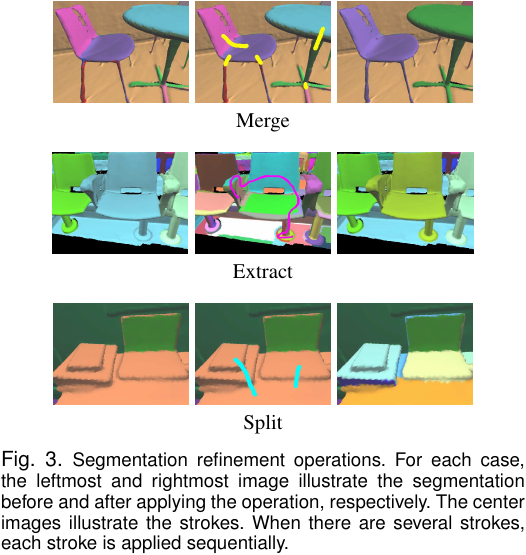}
    \caption{}
    \label{fig:segmentationrefinement}
\end{figure}

\textbf{Merge.}
This operation is used to resolve over-segmentation. In particular, users identify over-segmented regions that need to be grouped by stroking on them. The merge operation is illustrated in the first row of Fig.~\ref{fig:segmentationrefinement}.

\textbf{Extract.}
This operation is designed to handle under-segmentation. In particular, users first select an under-segmented region, the supervertices composing the under-segmented region are retrieved. Users then can select a few supervertices and use the merge operation to group those supervertices to create a new region. Note that the supervertices are not recomputed. Instead, they are retrieved from the cache result in the graph-based segmentation step. The second row of Fig.~\ref{fig:segmentationrefinement} shows the extract operation.

\textbf{Split.} In a few rare cases, the MRF-based segmentation may perform differently on different regions. This is probably because of the variation of the geometric shape and appearance of objects. For example, a scene may have chairs appearing in a unique color and other chairs each of which composes multiple colors. Therefore, a unique setting of the parameters in the MRF-based segmentation may not adapt to all objects.

\begin{figure*}[t]
    \centering
    \includegraphics[page=4]{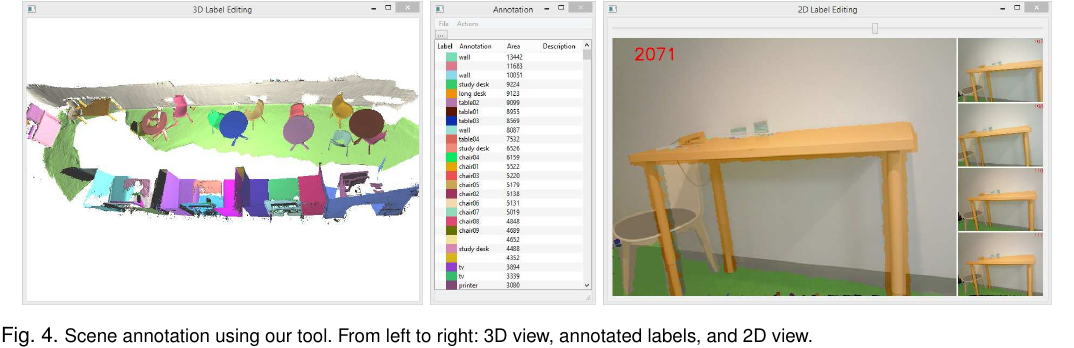}
    \caption{}
    \label{fig:annotation}
\end{figure*}

To address this issue, we design a split operation enabling user-guided MRF-based segmentation. Specifically, users first select an under-segmented region by stroking on that region. The MRF-based segmentation is then invoked on the selected region with a small value of $\gamma$ (see (\ref{eq:energy})) to generate more grained regions. We then enforce a constraint such that the starting and ending point of the stroke belong to two different regions. For example, assume that $l_i$ and $l_j$ are the labels of two supervertices that respectively contain the starting and ending point of the stroke. To bias the objective function in (\ref{eq:energy}), $\psi_2(l_i, l_j)$ in (\ref{eq:prior}) is set to $-1$ when $l_i \neq l_j$, and to a large value (e.g. $10^9$) otherwise. By doing so, the optimization in (\ref{eq:energy}) would favor the case $l_i \neq l_j$. In other words, the supervertices at the starting and ending point are driven to separate regions. Note that the MRF-based segmentation is only re-executed on the selected region. Therefore, the split operation is fast and does not hinder user interaction. The third row of Fig.~\ref{fig:segmentationrefinement} illustrates the split operation.

Through experiments we have found that most of the time, users perform merge and extract operations. Split operation is only used when extract operation is not able to handle severe under-segmentations but such cases are not common in practice. When all the 3D segmented regions have been refined, users can annotate the regions by providing the object type, e.g. coffee table, sofa chair. Fig.~\ref{fig:annotation} shows an example of using our tool for annotation. Note that users are free to navigate the scene in both 3D and 2D space.

\section{Object Search}
\label{sec:objectsearch}

There may exist multiple instances of an object class in a scene, e.g. the nine chairs in Fig.~\ref{fig:shapematching}. To support labeling and annotating repetitive objects, users can define a template by selecting an existing region or multiple regions composing the template. Those regions are the results of the MRF-based segmentation or user refinement. Given the user-defined template, our system automatically searches for objects that are similar to the template. Note that the repetitive objects are not present as a single region. Instead, each repetitive object may be composed of multiple regions. For example, each chair in Fig.~\ref{fig:shapematching}(a) consists of different regions such as the back, seat, legs. Once a group of regions is found to well match with the template, the regions are merged into a single object and recommended to users for verification. We extend the 2D shape context proposed in \cite{Belongie_PAMI_2002} to describe 3D objects (section~\ref{sec:shapecontext}). Matching objects with the template is performed via comparing shape context descriptors (section~\ref{sec:shapematching}). The object search is then built upon the sliding-window object detection approach \cite{Dalal_CVPR_2005} (section~\ref{sec:searching}).

\subsection{Shape Context}
\label{sec:shapecontext}

\begin{figure}[t]
    \centering
    \includegraphics[page=5]{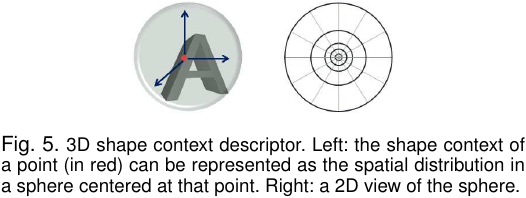}
    \caption{}
    \label{fig:3Dshapecontext}
\end{figure}

Shape context was proposed by Belongie et al. \cite{Belongie_PAMI_2002} as a 2D shape descriptor and is well-known for many desirable properties such as being discriminative, robust to shape deformation and transformation, and less sensitive to noise and partial occlusions. Those properties fit well our need for several reasons. First, reconstructed scene meshes could be incomplete and contain noisy surfaces. Second, occlusions may also appear due to the lack of sufficient images completely covering objects. Third, the tool is expected to adapt with the variation of object shapes, e.g. chairs with and without arms.


In our work, a 3D object is represented by a set $\mathcal{V}$ of vertices obtained from the 3D reconstruction step. For each vertex $\mathbf{v}_i \in \mathcal{V}$, the shape context of $\mathbf{v}_i$ is denoted as $\mathbf{s}(\mathbf{v}_i)$ and represented by the histogram of the relative locations of other vertices $\mathbf{v}_j$, $j \neq i$, to $\mathbf{v}_i$. Let $\mathbf{u}_{ij} = \mathbf{v}_i - \mathbf{v}_j$. The relative location of a vertex $\mathbf{v}_j \in \mathcal{V}$ to $\mathbf{v}_i$ is encoded by the length $\| \mathbf{u}_{ij} \|$ and the spherical coordinate $(\theta, \phi)_{ij}$ of $\mathbf{u}_{ij}$. In our implementation, the lengths $\| \mathbf{u}_{ij} \|$ were quantized into 5 levels. To make the shape context $\mathbf{s}(\mathbf{v}_i)$ more sensitive to local deformations, $\| \mathbf{u}_{ij} \|$ were quantized in a log-scale space. The spherical angles $(\theta, \phi)_{ij}$ were quantized uniformly into 6 discrete values. Fig.~\ref{fig:3Dshapecontext} illustrates the 3D shape context.

The shape context descriptor is endowed with scale-invariant by normalizing $\| \mathbf{u}_{ij} \|$ by the mean of the lengths of all vectors. To make the shape context rotation invariant, Kortgen et al. \cite{Kortgen_ESCG_2003} computed the spherical coordinates $(\theta, \phi)_{ij}$ relatively to the eigenvectors of the covariance matrix of all vertices. However, the eigenvectors may not be computed reliably for shapes having no dominant orientations, e.g. rounded objects. In addition, the eigenvectors are only informative when the shape is complete while our scene meshes may be incomplete. To overcome this issue, we establish a local coordinate frame at each vertex on a shape using its normal and tangent vector. The tangent vector of a vertex $\mathbf{v}_i$ is the one connecting $\mathbf{v}_i$ to the centroid of the shape. We have found this approach worked more reliably.

Since a reconstructed scene often contains millions of vertices, prior to applying the object search, we uniformly sample a scene by $20,000$ points which result in objects of $100-200$ vertices.

\subsection{Shape Matching}
\label{sec:shapematching}

Comparing (matching) two given shapes $\mathcal{V}$ and $\mathcal{Y}$ is to maximize the correspondences between pairs of vertices on these two shapes, i.e. minimizing the deformation of the two shapes in a point-wise fashion. The deformation cost between two vertices $\mathbf{v}_i \in \mathcal{V}$ and $\mathbf{y}_j \in \mathcal{Y}$ is measured by the $\chi^2(\mathbf{s}(\mathbf{v}_i), \mathbf{s}(\mathbf{y}_j))$ distance between the two corresponding shape context descriptors extracted at $\mathbf{v}_i$ and $\mathbf{y}_j$ as follow,
\begin{align}
  \label{eq:chisquareddistance}
  \chi^2(\mathbf{s}(\mathbf{v}_i), \mathbf{s}(\mathbf{y}_j)) = \frac{1}{2}\sum_{b=1}^{\dim(\mathbf{s}(\mathbf{v}_i))} \frac{(\mathbf{s}(\mathbf{v}_i)[b]-\mathbf{s}(\mathbf{y}_j)[b])^2}{\mathbf{s}(\mathbf{v}_i)[b]+\mathbf{s}(\mathbf{y}_j)[b]}
\end{align}
where $\dim(\mathbf{s}(\mathbf{v}_i))$ is the dimension (i.e. the number of bins) of $\mathbf{s}(\mathbf{v}_i)$, $\mathbf{s}(\mathbf{v}_i)[b]$ is the value of $\mathbf{s}(\mathbf{v}_i)$ at the $b$-th bin.

Given the deformation cost of every pair of vertices on two shapes $\mathcal{V}$ and $\mathcal{Y}$, shape matching can be solved using the shortest augmenting path algorithm \cite{Jonker_Computing_1987}. To make the matching algorithm adaptive to shapes with different number of vertices, ``dummy'' vertices are added. This enables the matching method to be robust to noisy data and partial occlusions. Formally, the deformation cost $C(\mathcal{V},\mathcal{Y})$ between two shapes $\mathcal{V}$ and $\mathcal{Y}$ is computed as,
\begin{equation}
\label{eq:deformationcost}
    C(\mathcal{V},\mathcal{Y}) = \sum_{\mathbf{v}_i \in \hat{\mathcal{V}}} \chi^2(\mathbf{s}(\mathbf{v}_i), \mathbf{s}(\pi(\mathbf{v}_i)))
\end{equation}
where $\hat{\mathcal{V}}$ is identical to $\mathcal{V}$ or augmented from $\mathcal{V}$ by adding dummy vertices and $\pi(\mathbf{v}_i) \in \hat{\mathcal{Y}}$ is the matching vertex of $\mathbf{v}_i$ determined by using \cite{Jonker_Computing_1987}.

To further improve the matching, we also consider how well the two matching shapes are aligned. In particular, we first align $\mathcal{V}$ to $\mathcal{Y}$ using a rigid transformation. This rigid transformation is represented by a $4 \times 4$ matrix and estimated using the RANSAC algorithm that randomly picks three pairs of correspondences and determine the rotation and translation \cite{Horn_JOSA_1987}. We then compute an alignment error,
\begin{equation}
\label{eq:alignmenterror}
    E(\mathcal{V}, \mathcal{Y}) = \min \bigg\{ \sqrt{ \frac{1}{|\mathcal{V}|} \sum_{i=1}^{|\mathcal{V}|} \epsilon^{(\mathcal{V})}_i}, \sqrt{ \frac{1}{|\mathcal{Y}|} \sum_{i=1}^{|\mathcal{Y}|} \epsilon^{(\mathcal{Y})}_i} \bigg\}
\end{equation}
where
\begin{equation}
\epsilon^{(\mathcal{V})}_i =
\begin{cases}
 \| \pi(\mathbf{v}_i) - T * \mathbf{v}_i \|^2	&	\text{if } \pi(\mathbf{v}_i) \text{ exists for } \mathbf{v}_i \in \mathcal{V} \\
 \Delta^2 & \text{otherwise}
\end{cases}
\end{equation}
and, similarly for $\epsilon^{(\mathcal{Y})}_i$, where $T$ is the rigid transformation matrix and $\Delta$ is a large value used to penalize misalignments.

A match is confirmed if: (i) $C(\mathcal{V}, \mathcal{Y}) < \tau_s$ and (ii) $E(\mathcal{V}, \mathcal{Y}) < \tau_a$ where $\tau_s$ and $\tau_a$ are thresholds. In our experiments, we set $\Delta = 2$ (meters), $\tau_s = 0.7$, $\tau_a = 0.4$. We have found that the object search method was not too sensitive to parameter settings while those settings achieved the best performance.

\begin{figure*}
    \centering
    \includegraphics[page=6]{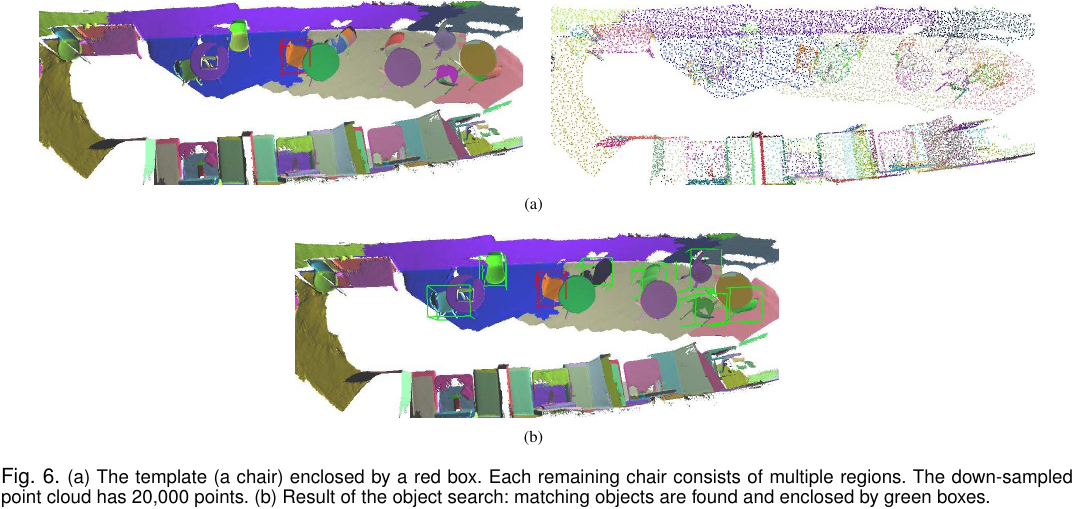}
    \caption{}
    \label{fig:shapematching}
\end{figure*}

\subsection{Searching}
\label{sec:searching}

Object search can be performed based on the sliding-window approach \cite{Dalal_CVPR_2005}. Specifically, we take the 3D bounding box of the template and use it as the window to scan a 3D scene. At each location in the scene, all regions that intersect the window are considered for their possibility to be part of a matching object. However, it would be intractable to consider every possible combination of all regions. To deal with this issue, we propose a greedy algorithm that operates iteratively by adding and removing regions.

\begin{algorithm}
    \caption{Grow-shrink procedure. $C$ and $E$ are the matching cost and alignment error defined in (\ref{eq:deformationcost}) and (\ref{eq:alignmenterror}).}
    \label{alg:findingobjs}
    \SetKwInOut{Input}{Input}
    \SetKwInOut{Output}{Output}
    \SetKwFunction{PopUp}{PopUp}
    \SetKwData{True}{true}
    \SetKwData{False}{false}
	\SetKw{And}{and}
    \underline{function GrowShrink} $(\mathcal{R},\mathcal{W}, \mathcal{O})$\\
    \Input{$\mathcal{R}$: set of regions to examine, \\
           $\mathcal{W}$: window,\\
           $\mathcal{O}$: user-defined template}
    \Output{$\mathcal{A}$: best matching object}
    \Begin{
        $\mathcal{A} \leftarrow \mathcal{R}$\

		\For{$i\leftarrow 1$ \KwTo $iterations$}
        {
        	\tcp{grow}	
            $M \leftarrow A$\

            \For{$r \in \mathcal{W} \setminus \mathcal{M}$}
            {
            	\If{$C(\mathcal{M} \cup \{r\}, \mathcal{O}) < C(\mathcal{A}, \mathcal{O})$ \And $E(\mathcal{M} \cup \{r\}, \mathcal{O}) < \tau_a$}
	            {
                    $\mathcal{A} \leftarrow \mathcal{M} \cup \{r\}$\
    	        }
			}

            \tcp{shrink}
            $M \leftarrow A$\

            \For{$r \in \mathcal{M}$}
            {
	            \If{$C(\mathcal{M} \setminus \{r\}, \mathcal{O}) < C(\mathcal{A}, \mathcal{O})$ \And $E(\mathcal{M} \setminus \{r\}, \mathcal{O}) < \tau_a$}
    	        {
                 	$\mathcal{A} \leftarrow \mathcal{M} \setminus \{r\}$\
        	    }
            }
        }

        \Return{$\mathcal{A}$}
    }
\end{algorithm}

The general idea is as follows. Let $\mathcal{R}$ be the set of regions that intersect the window $\mathcal{W}$, i.e. the 3D bounding box of the template. For a region $r \in \mathcal{W} \setminus \mathcal{R}$, we verify whether the object made by $\mathcal{R} \cup \{r\}$ could be more similar to the user-defined template $\mathcal{O}$ in comparison with $\mathcal{R}$. Similarly, for every region $r \in \mathcal{R}$ we also verify the object made by $\mathcal{R} \setminus \{r\}$. These adding and removing steps are performed interchangeably in a small number of iterations until the best matching result (i.e. a group of regions) is found. This procedure is called \emph{grow-shrink} and described in Algorithm~\ref{alg:findingobjs}.

In our implementation, the spatial strides on the $x-$, $y-$, and $z-$ direction of the window $\mathcal{W}$ were set to the size of $\mathcal{W}$. The number of iterations in Algorithm~\ref{alg:findingobjs} was set to $10$, which resulted in satisfactory accuracy and efficiency.

Since a region may be contained in more than one window, it may be verified multiple times in multiple groups of regions. To avoid this, if an object candidate is found in a window, its regions will not be considered in any other objects and any other windows. Fig.~\ref{fig:shapematching} illustrates the robustness of the object search in localizing repetitive objects under severe conditions (e.g. objects with incomplete shape).

\begin{figure}
    \centering
    \includegraphics[page=8]{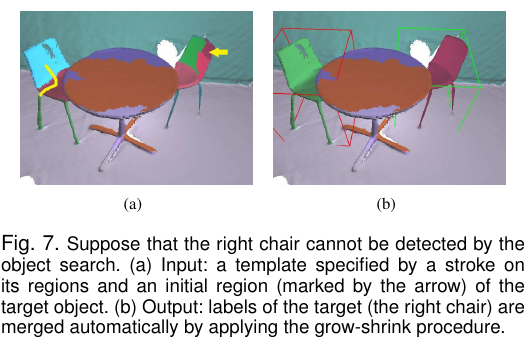}
    \caption{}
    \label{fig:guidedmerge}
\end{figure}

The search procedure may miss some objects. To handle such cases, we design an operation called \emph{guided merge}. In particular, after defining the template, users simply select one of the regions of a target object that is missed by the object search. The grow-shrink procedure is then applied on the selected region to seek a better match with the template. Fig.~\ref{fig:guidedmerge} shows an example of the guided merge operation.

\section{Segmentation on 2D}
\label{sec:2Dsegmentation}

\begin{figure*}
    \centering
    \includegraphics[page=9]{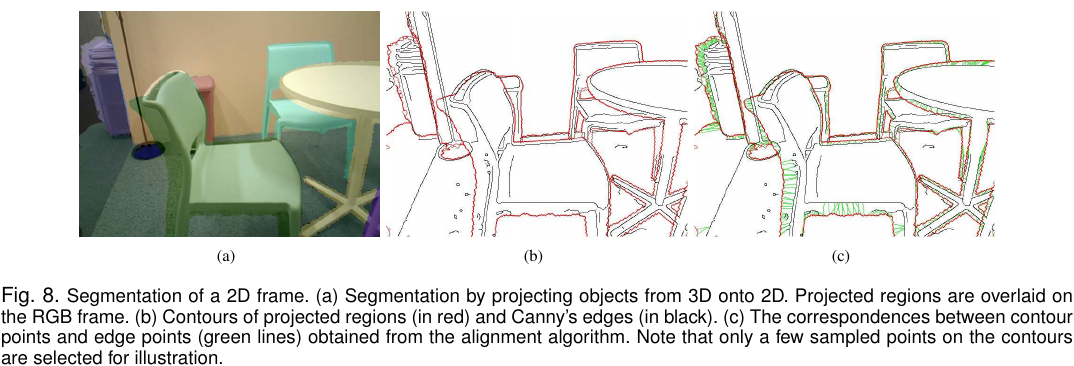}
    \caption{}
    \label{fig:2Dseginterim}
\end{figure*}

Segmentation on 2D can be done by projecting regions in 3D space onto 2D frames. However, the projected regions may not well align with the true objects on 2D frames (see Fig.~\ref{fig:2Dseginterim}). There are several reasons for this issue. For example, the depth and color images used to reconstruct a scene might not be exactly aligned at object boundaries; the camera intrinsics are from factory settings and not well calibrated; camera registration during reconstruction exhibits drift.

\begin{figure*}
    \centering
    \includegraphics[page=10]{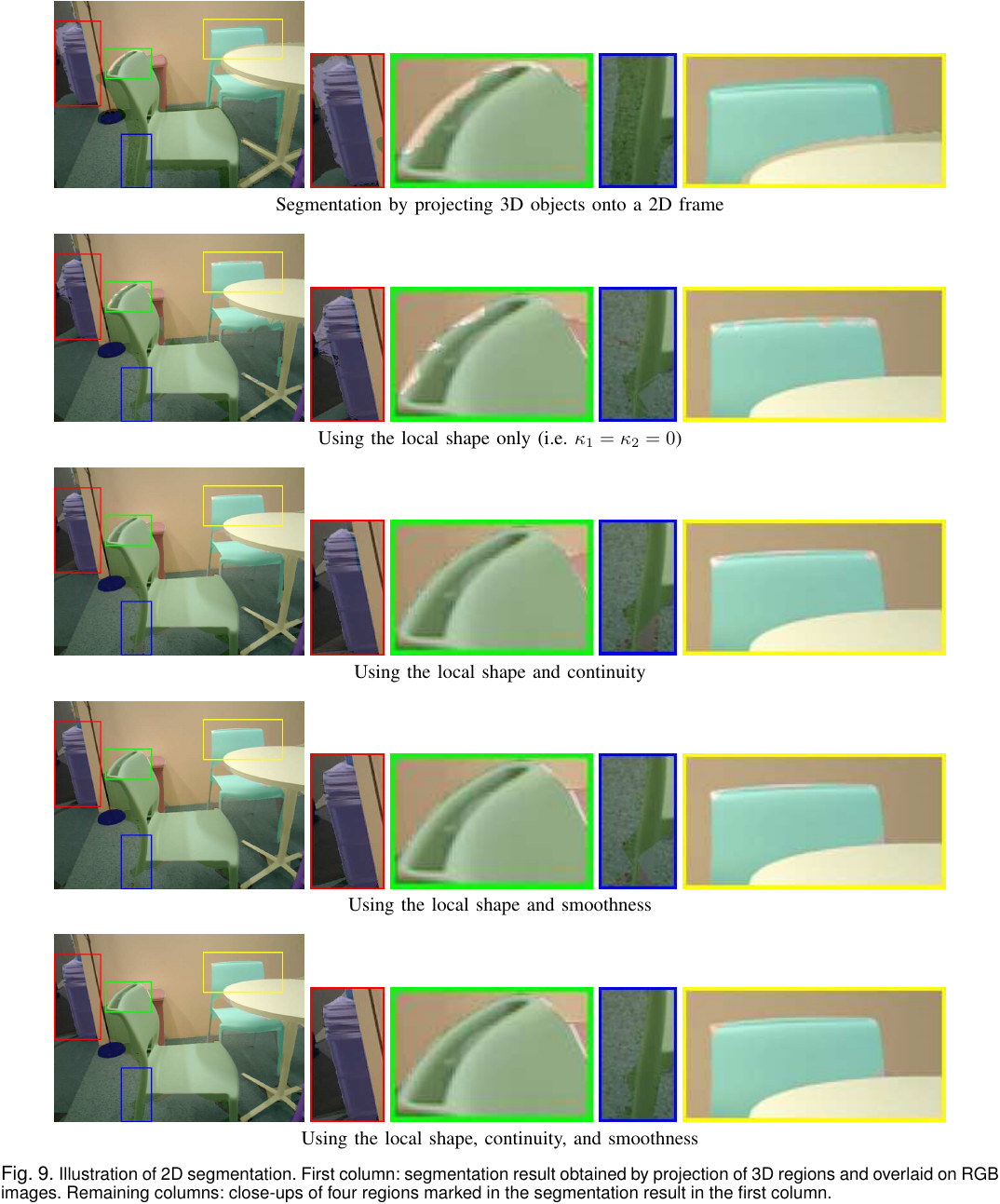}
    \caption{}
    \label{fig:misalignment}
\end{figure*}

To overcome this issue, we propose an alignment algorithm which aims to fit the boundaries of projected regions to true boundaries on 2D frames. The true boundaries on a 2D frame can be extracted using some edge detector (e.g. the Canny edge detector \cite{Canny_PAMI_1986}). Let $E=\{e_j\}$ denote the set of edge points on the edge map of a 2D frame. Let $U=\{u_i\}$ be the set of contour points of a projected object on that frame. $U$ is then ordered using the Moore neighbor tracing algorithm \cite{Narappanawar_CVIU_2011}. The ordering step is used to express the contour alignment problem in a form that dynamic programming can be applied for efficient implementation.

At each contour point $u_i$, we consider a $21 \times 21$-pixel window centered at $u_i$ (in relative to a $640 \times 480$-pixel image). We then extract the histogram $h_{u_i}$ of the orientations of vectors $(u_i,u_k)$, $u_k \in U$ in the window. The orientations are uniformly quantized into 16 bins. We also perform this operation for edge points $e_j \in E$. The dissimilarity between the two local shapes at a contour point $u_i$ and edge point $e_j$ is computed as $\chi^2(h_{u_i}, h_{e_j})$ (similarly to (\ref{eq:chisquareddistance})).


We also consider the continuity and smoothness of contours. In particular, the continuity between two adjacent points $u_{i}$ and $u_{i-1}$ is defined as $\|u_{i} - u_{i-1}\|$. The smoothness of a fragment including three consecutive points $u_i$, $u_{i-1}$, $u_{i-2}$ is computed as $\cos(u_i - u_{i-1}, u_{i-2}-u_{i-1})$ where $u_i - u_{i-1}$ and $u_{i-2}-u_{i-1}$ denote the vectors connecting $u_{i-1}$ to $u_i$ and connecting $u_{i-1}$ to $u_{i-2}$ respectively, and $\cos(\cdot, \cdot)$ is the cosine of the angle formed by these two vectors.

Alignment of $U$ to $E$ is to identify a mapping function $f: U \rightarrow E$ that maps a contour point $u_i \in U$ to an edge point $e_j \in E$ so as to,
\begin{align}
  \label{eq:alignmentobjective}
  &\mbox{minimize} \bigg[ \sum_{i=1}^{|U|} \chi^2(h_{u_i}, h_{f(u_i)}) \notag \\
   &+ \kappa_1 \sum_{i=2}^{|U|} \|f(u_{i}) - f(u_{i-1})\| \notag \\
   &+ \kappa_2 \sum_{i=3}^{|U|} \cos(f(u_i) - f(u_{i-1}), f(u_{i-2})-f(u_{i-1})) \bigg]
\end{align}

The optimization problem in (\ref{eq:alignmentobjective}) can be considered as the bipartite graph matching problem \cite{Jonker_Computing_1987}. However, since $U$ is ordered, this optimization can be solved efficiently using dynamic programming \cite{Thayananthan_CVPR_2003}. In particular, denoting $m_{i,j}=\chi^2(h_{u_i}, h_{e_j})$, $f_i=f(u_i)$, $f_{i,j}=f(u_i)-f(u_j)$, the objective function in (\ref{eq:alignmentobjective}) can be rewritten as,
\begin{align}
  \label{eq:alignmentdynamicprogramming}
  \mathcal{F}_i &= \begin{cases}
    \min_{j \in E} \{\mathcal{F}_{i-1} + m_{i,j} + \kappa_1 \|f_{i,i-1}\| \\ + \kappa_2 \cos(f_{i,i-1},f_{i-2,i-1})\}, & \mbox{if }i>2\\
    \min_{j \in E} \{\mathcal{F}_{i-1} + m_{i,j} + \kappa_1 \|f_{i,i-1}\| \}, & \mbox{if }i=2\\
    \min_{j \in E} \{m_{i,j}\}, & \mbox{if }i=1
    \end{cases}
\end{align}
where $\kappa_1$ and $\kappa_2$ are user parameters. We have tried $\kappa_1$ and $\kappa_2$ with various values and found that $\kappa_1=0.1$ and $\kappa_2=3.0$ often produced good results.

In (\ref{eq:alignmentdynamicprogramming}), for each contour point $u_i$, all edge points are verified for a match. However, this exhausted search is not necessary since the misalignment only occurs at a certain amount. To save the computational cost, we limit the search space for each contour point $u_i$ by considering only its $k$ nearest edge points whose the distance to $u_i$ is less than a distance $\delta$. In our experiment, $\delta$ was set to $10\%$ of the maximum of the image dimension, e.g., $\delta=48$ for a $640\times480$-pixel image. The number of nearest edge points (i.e. $k$) was set to 30. Fig.~\ref{fig:2Dseginterim}(c) shows an example of contour alignment by optimizing (\ref{eq:alignmentdynamicprogramming}) using dynamic programming.

We have also verified the contribution of the continuity and smoothness. Fig.~\ref{fig:misalignment} shows the results when the cues are used individually and in combination. The results show that, when all the cues are taken into account, the contours are mostly well aligned with the true object boundaries. It is noticed that the seat of the green chair is not correctly recovered. We have found that this is because the Canny's detector missed important edges on the boundaries of the chair. Users are also free to edit the alignment results.

\section{Experiments}
\label{sec:experiments}

We present the dataset on which experiments were conducted in section~\ref{sec:data}. We evaluate the 3D segmentation in section~\ref{sec:3Dsegmentationevaluation}. The object search is evaluated in section~\ref{sec:objectsearchevaluation}. Experimental results of the 2D segmentation are finally presented in section~\ref{sec:2Dsegmentationevaluation}.

\subsection{Dataset}
\label{sec:data}


\begin{figure*}
    \centering
    \includegraphics[page=11]{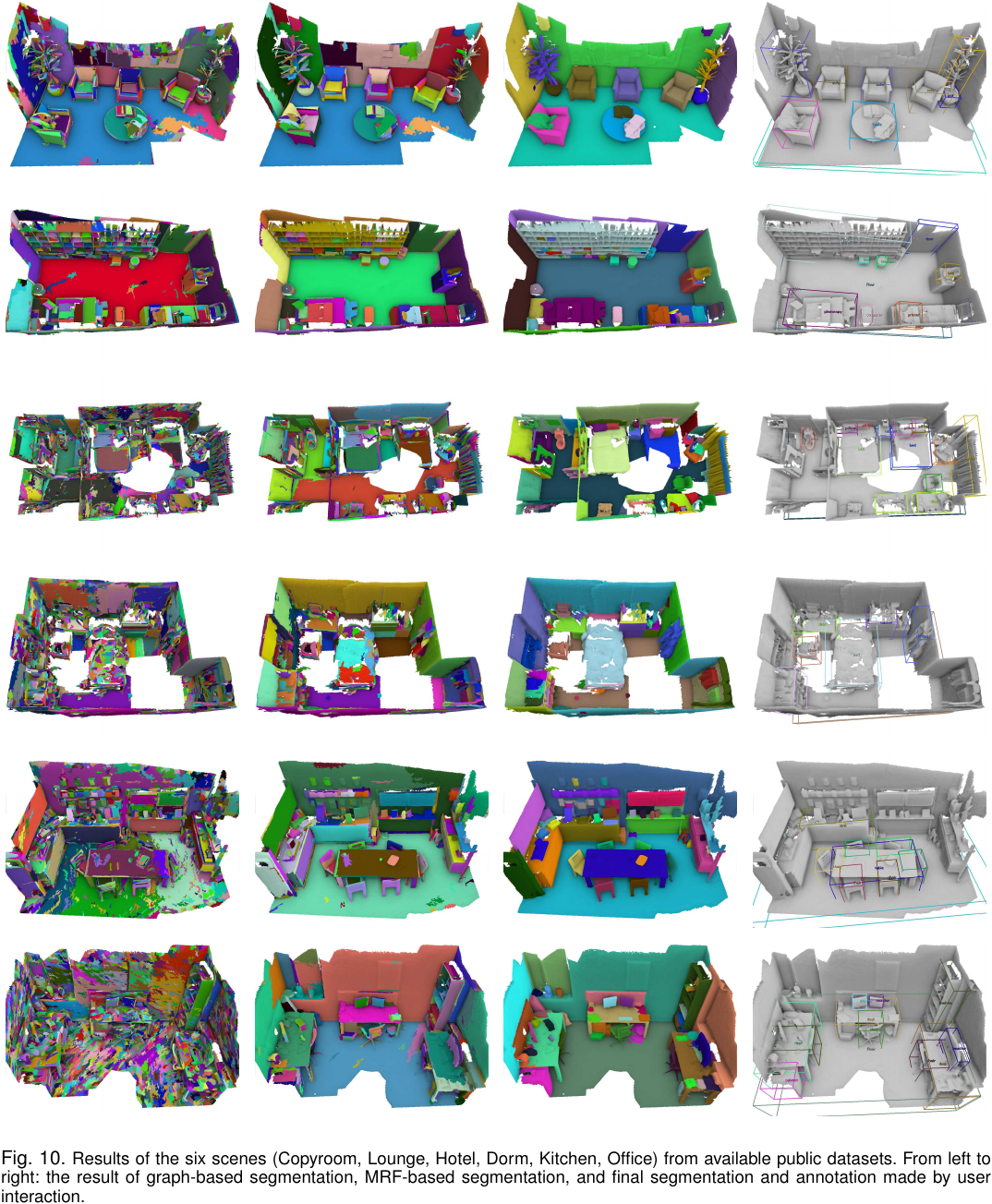}
    \caption{}
    \label{fig:results1}
\end{figure*}

\begin{figure*}
    \centering
    \includegraphics[page=12]{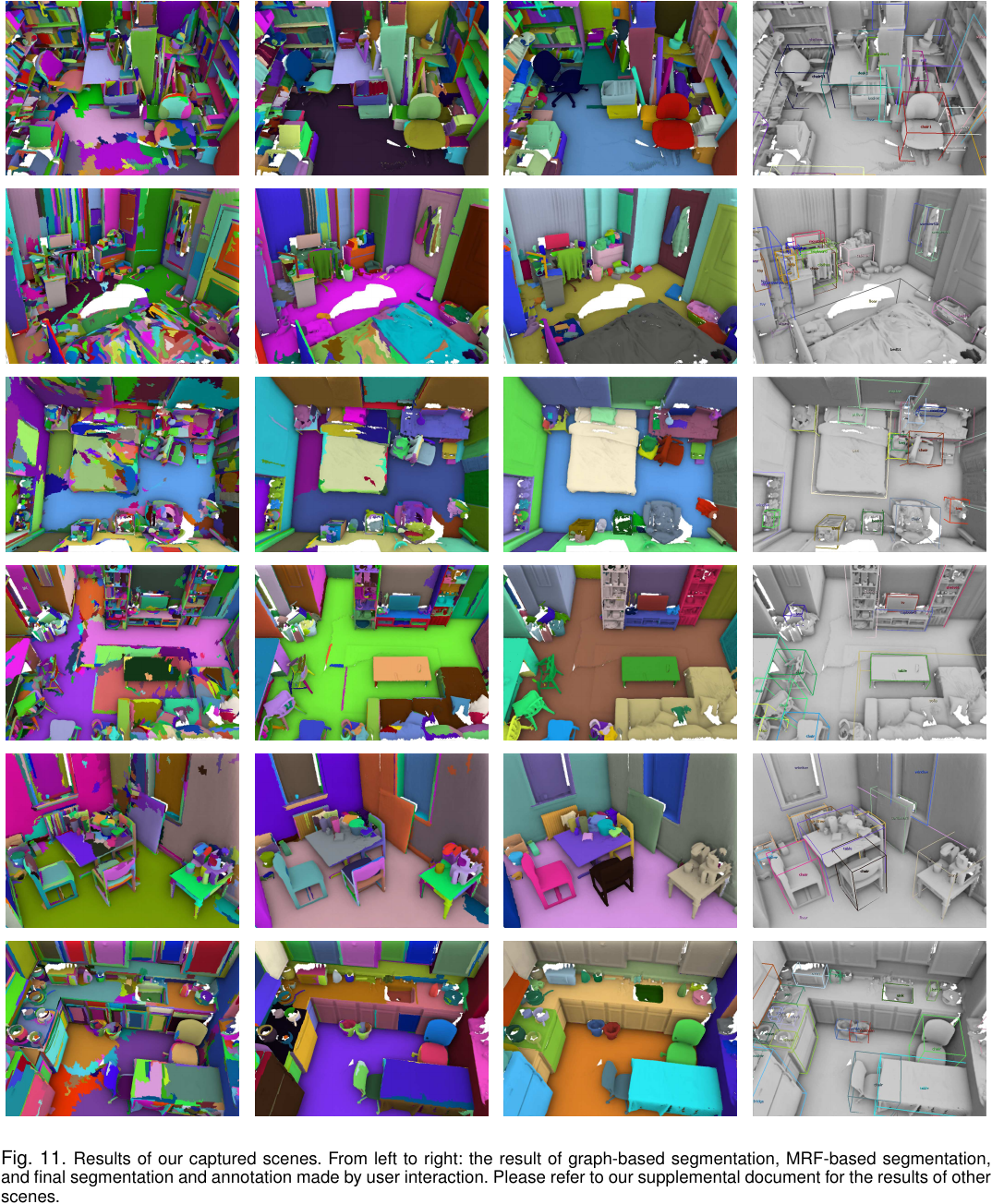}
    \caption{}
    \label{fig:results2}
\end{figure*}

\begin{table*}[!ht]
\caption{\small Comparison of the graph-based segmentation and MRF-based segmentation. For our captured scenes, the statistical data is the average numbers calculated over all the scenes. Note that for user refined results, the numbers of objects annotated are fewer than the numbers of labels (i.e. segments). This is because the annotation was done only for objects that are common in practice.}
\label{tab:3DSeg}
\centering
\small
\begin{tabular}{lrrrrrrrr}
\toprule
& & \multicolumn{2}{c}{Graph-based} & \multicolumn{2}{c}{MRF-based} &  \multicolumn{2}{c}{User refined} & Interactive time\\
\cmidrule{3-8}
Scene & \#Vertices & \#Supervertices & OCE  & \#Regions & OCE & \#Labels & \#Objects & (in minutes) \\
\midrule
\emph{copyroom} & 1,309,421 & 1,996 & 0.92 & 347 & 0.73 & 157 & 15 & 19 \\
\emph{lounge} & 1,597,553 & 2,554 & 0.97 & 506 & 0.93 & 53 & 12 & 16 \\
\emph{hotel} & 3,572,776 & 13,839 & 0.98 & 1433 & 0.88 & 96 & 21 & 27 \\
\emph{dorm} & 1,823,483 & 3,276 & 0.97 & 363 & 0.78 & 75 & 10 & 15 \\
\emph{kitchen} & 2,557,593 & 4,640 & 0.97 & 470 & 0.85 & 75 & 24 & 23 \\
\emph{office} & 2,349,679 & 4,026 & 0.97 & 422 & 0.84 & 69 & 19 & 24 \\
\midrule
Our scenes & 1,450,748 & 2,498 & 0.93 & 481 & 0.77 & 179 & 19 & 30 \\
\bottomrule
\end{tabular}
\end{table*}

We created a dataset consisting of over 100 scenes. The dataset includes six scenes from publicly available datasets: the \textit{copyroom} and \textit{lounge} from the Stanford dataset \cite{Zhou_TOG_2013}, the \textit{hotel} and \textit{dorm}
from the SUN3D \cite{Xiao_ICCV_2013}, and the \textit{kitchen} and \textit{office} sequences from the Microsoft dataset \cite{Shotton_CVPR_2013}. The Stanford and SUN3D dataset also provide registered RGB and depth image pairs. These datasets also include the camera pose data.

In addition to existing scenes, we collected 100 scenes using Asus Xtion and Microsoft Kinect v2. Our scenes were captured from the campus of the University of Massachusetts Boston and the Singapore University of Technology and Design. These scenes were captured from various locations such as lecturer rooms, theatres, university hall, library, computer labs, dormitory, etc. All the scenes were then fully segmented and annotated using our tool. The dataset also includes the camera pose information. Fig.~\ref{fig:results1} and Fig.~\ref{fig:results2} show the six scenes collected from the public datasets and several of our collected scenes.

\subsection{Evaluation of 3D Segmentation}
\label{sec:3Dsegmentationevaluation}

We evaluated the impact of the graph-based and MRF-based segmentation on our dataset. We considered the annotated results obtained using our tool as the ground-truth. To measure the segmentation performance, we extended the object-level consistency error (OCE), the image segmentation evaluation metric proposed in \cite{Polak_IVC_2009} to 3D vertices. Essentially, the OCE reflects the coincidence of pixels/vertices of segmented regions and ground-truth regions. As indicated in \cite{Polak_IVC_2009}, compared with other segmentation evaluation metrics (e.g. the global and local consistency error in \cite{Martin_ICCV_2001}), the OCE considers both over- and under-segmentation errors in a single measure. In addition, OCE can quantify the accuracy of multi-object segmentation and thus it fits well our evaluation purpose.


Table~\ref{tab:3DSeg} summarizes the OCE of the graph-based and MRF-based segmentation. As shown in the table, compared with the graph-based segmentation, the segmentation accuracy is significantly improved by the MRF-based segmentation. It is also noticeable on the reduction of the quantity of the 3D vertices to the supervertices and the regions. However, experimental results also show that the segmentation results generated automatically are still not approaching the quality made by human beings. Thus, user interactions are necessary. This is because of two reasons. First, both the graph-based and MRF-based segmentation aim to segment a 3D scene into homogenous regions/surfaces rather than semantical objects. Second, the semantic segmentation done by users are subjective. For example, one may consider a pot and a plant growing on it as two separate objects or as a single object.

After user interaction, the number of final labels are typically less than a hundred. The number of semantic objects is around 10 to 20 in most of the cases. Note that the numbers of final labels and semantic objects are not identical. This is because there could have labels whose semantics is not well defined, e.g. miscellaneous items on a table or some small labels appeared as noise in the 3D reconstruction.

We also measured the time required for user interactions using our tool. This information is reported in the last column of Table~\ref{tab:3DSeg}. As shown in the table, with the assistance of the tool, complex 3D scenes (with millions of vertices) could be completely segmented and annotated in less than 30 minutes, as opposed to approximately few hours to be done manually. Note that the interactive time is subjective to user's experience. Several results of our tool on the public datasets and our collected dataset are shown in Fig.~\ref{fig:results1} and Fig.~\ref{fig:results2}.

Through experiments we have found that although our tool was able to work with most reconstructed scenes in reasonable processing time, it failed at a few locally rough terrains, e.g. the 3D mesh outer boundaries and the pothole areas made by loop closure. Enhancing broken surfaces and completing missing object parts will be our future work.

\subsection{Evaluation of Object Search}
\label{sec:objectsearchevaluation}

To evaluate the object search functionality, we collected a set of 45 objects from our dataset. Those objects were selected so that they are semantical and common in practice and their shapes are discriminative. For example, drawers of cabinets were not selected since they were present in flat surfaces which could be easily found in many structures, e.g. walls, pictures, etc.  For each scene and each object class (e.g. chair), each object in the class was used as the template while the remaining repetitive objects of the same class were considered as the ground-truth. The object search was then applied to find repetitive objects given the template.


We used the precision, recall, and $F$-measure to evaluate the performance of the object search. The intersection over union (IoU) metric proposed in \cite{Everingham_IJCV_2010} for object detection was used as the criterion to determine true detections and false alarms. However, instead of computing the IoU on the bounding boxes of objects as in \cite{Everingham_IJCV_2010}, we entailed the IoU at point-level (i.e. 3D vertices from the mesh). This is because our aim is not only to localize repetitive objects but also to segment them. In particular, an object $\mathcal{O}$ (a set of vertices) formed by the object search procedure is considered as true detection if there exists an annotated object $\mathcal{R}$ in the ground-truth such that
\begin{equation}
\frac{|\mathcal{O} \cap \mathcal{R}|}{|\mathcal{O} \cup \mathcal{R}|} > 0.5
\end{equation}
where $|\cdot|$ denotes the area; the value 0.5 is often used in object detection evaluation (e.g. \cite{Everingham_IJCV_2010}).

The evaluation was performed on every template. The precision, recall, and $F$-measure ($=2\times\frac{Precision\times Recall}{Precision+Recall}$) were then averaged over all evaluations. Table~\ref{tab:objectsearch} shows the averaged precision, recall, and $F$-measure of the object search. As shown, the tool can localize and segment 70\% of repetitive objects with 69\% precision and 65\% $F$-measure. We also tested the object search without considering the alignment error (i.e. $E$ in \ref{eq:alignmenterror})). Experimental results show that, compared with the solely use of shape context dissimilarity score (i.e. $C$ in (\ref{eq:deformationcost})), while the augmentation of alignment error could slightly incur a loss of the detection rate (about 2\%), it largely improved the precision (from 22\% to 69\%). This led to a significant increase of the $F$-measure (from 30\% to 65\%).

Our experimental results show that, the object search worked efficiently with templates represented by about 200 points. For example, for the scene presented in Fig.~\ref{fig:shapematching}, the object search was completed within 15 seconds with a 150-point template and on a machine equipped by an Intel(R) Core(TM) i7 2.10 GHz CPU and 32 GB of memory. In practice, threads can be used to run the object search in the background while users are performing interactions.

\begin{table}[!ht]
\caption{\small Performance of the proposed object search.}
\label{tab:objectsearch}
\centering
\small
\begin{tabular}{lccc}
\toprule
& Precision & Recall & $F$-measure \\
\midrule
Without alignment error & 0.22 & \textbf{0.72} & 0.30 \\
With alignment error & \textbf{0.69} & 0.70 & \textbf{0.65} \\
\bottomrule
\end{tabular}
\end{table}

\subsection{Evaluation of 2D Segmentation}
\label{sec:2Dsegmentationevaluation}

We also evaluated the performance of the segmentation on 2D using the OCE metric. This experiment was conducted on the \textit{dorm} sequence from the SUN3D dataset \cite{Xiao_ICCV_2013}. The \textit{dorm} sequence contained 58 images in which the ground-truth labels were manually crafted and publicly available.


We report the segmentation performance obtained by projecting 3D regions onto 2D images and by applying the our alignment algorithm in Table~\ref{tab:2DSeg}. The impact of the local shape, continuity, and smoothness is also quantified. As shown in Table~\ref{tab:2DSeg}, the combination of the local shape, continuity, and smoothness achieves the best performance. We have visually found the alignment algorithm could make projected contours smoother and closer to true edges and this would be more convenient for users to edit 2D segmentation results.

Experimental results show our alignment algorithm worked efficiently. On the average, the alignment could be done in about 1 second for a $640 \times 480$-pixel frame.


\begin{table}[!ht]
\caption{\small Comparison of different segmentation methods.}
\label{tab:2DSeg}
\centering
\small
\begin{tabular}{lc}
\toprule
Segmentation method & OCE \\
\midrule
Projection & 0.57 \\
Local shape & 0.60 \\
Local shape + Continuity & 0.55 \\
Local shape + Smoothness & 0.55 \\
Local shape + Continuity + Smoothness & \textbf{0.54} \\
\bottomrule
\end{tabular}
\end{table}

\section{Conclusion}
\label{sec:conclusion}

This paper proposed a robust tool for segmentation and annotation of 3D scenes. The tool couples the geometric information from 3D space and color information from multi-view 2D images in an interactive framework. To enhance the usability of the tool, we developed assistive user-interactive operations that allow users to flexibly manipulate scenes and objects in both 3D and 2D space. The tool is also facilitated with automated functionalities such as scene and image segmentation, object search for semantic annotation.

Along with the tool, we created a dataset of more than 100 scenes. All the scenes were annotated using our tool. The newly created dataset was also used to verify the tool. The overall performance of the tool depends on the quality of 3D reconstruction. Improving the quality of 3D meshes by recovering broken surfaces and missing object parts will be our future work.

\section*{Acknowledgment}

Lap-Fai Yu is supported by the University of Massachusetts Boston StartUp Grant P20150000029280 and by the Joseph P. Healey Research Grant Program provided by the Office of the Vice Provost for Research and Strategic Initiatives \& Dean of Graduate Studies of the University of Massachusetts Boston. This research is supported by the National Science Foundation under award number 1565978. We also acknowledge NVIDIA Corporation for graphics card donation.

Sai-Kit Yeung is supported by Singapore MOE Academic Research Fund MOE2013-T2-1-159 and SUTD-MIT International Design Center Grant IDG31300106. We acknowledge the support of the SUTD Digital Manufacturing and Design (DManD) Centre which is supported by the National Research Foundation (NRF) of Singapore. This research is also supported by the National Research Foundation, Prime Minister's Office, Singapore under its IDM Futures Funding Initiative.

Finally, we slenderly thank Fangyu Lin for assisting data capture and Guoxuan Zhang for the early version of the tool. 

\ifCLASSOPTIONcaptionsoff
  \newpage
\fi




\bibliographystyle{IEEEtran}
\bibliography{references}

\end{document}